\documentclass[10pt,twocolumn,letterpaper]{article}

\usepackage{iccv}              

\definecolor{iccvblue}{rgb}{0.21,0.49,0.74}
\usepackage[pagebackref,breaklinks,colorlinks,allcolors=iccvblue]{hyperref}

\def\paperID{8} 
\def\confName{ICCV}
\def\confYear{2025}

\title{Agro-Consensus: Semantic Self-Consistency in Vision-Language Models for Crop Disease Management in Developing Countries}

\author{Mihir Gupta\\
The Harker School, USA \\
{\tt\small mihirgupta292@gmail.com}
\and
Pratik Desai \\
Kissan.ai, USA \\
{\tt\small pratikDesai@outlook.com}
\and
Ross Greer\\
University of California, Merced, USA\\
{\tt\small rossgreer@ucmerced.edu}
}

\begin{document}
\maketitle

\begin{abstract}
Agricultural disease management in developing countries such as India, Kenya, and Nigeria faces significant challenges due to limited access to expert plant pathologists, unreliable internet connectivity, and cost constraints that hinder the deployment of large-scale AI systems. This work introduces a cost-effective self-consistency framework to improve vision-language model (VLM) reliability for agricultural image captioning. The proposed method employs semantic clustering, using a lightweight (80MB) pre-trained embedding model to group multiple candidate responses. It then selects the most coherent caption—containing a diagnosis, symptoms, analysis, treatment, and prevention recommendations—through a cosine similarity-based consensus. A practical human-in-the-loop (HITL) component is incorporated, wherein user confirmation of the crop type filters erroneous generations, ensuring higher-quality input for the consensus mechanism. Applied to the publicly available PlantVillage dataset using a fine-tuned 3B-parameter PaliGemma model, our framework demonstrates improvements over standard decoding methods. Evaluated on 800 crop disease images with up to 21 generations per image, our single-cluster consensus method achieves a peak accuracy of 83.1\% with 10 candidate generations, compared to the 77.5\% baseline accuracy of greedy decoding. The framework's effectiveness is further demonstrated when considering multiple clusters; accuracy rises to 94.0\% when a correct response is found within any of the top four candidate clusters, outperforming the 88.5\% achieved by a top-4 selection from the baseline.
\end{abstract}


\vspace{-2mm}
\section{Introduction}
\label{sec:intro}

For smallholder farmers in developing countries, timely and accurate crop disease diagnosis is a significant challenge that directly impacts food security and livelihoods. Access to expert plant pathologists is often limited, and unreliable internet connectivity hinders the adoption of many cloud-based agricultural technologies, creating an "expert gap" that can lead to significant crop losses.

Vision-language models (VLMs) offer a promising approach to address this deficit by providing automated, expert-level guidance directly to farmers. However, the performance of general-purpose VLMs can degrade in performance when applied to specialized agricultural domains \cite{10222401}. Furthermore, many existing systems are computationally intensive and require online connectivity, making them impractical for deployment in resource-constrained environments.

This work addresses these limitations through a self-consistency framework specifically designed to improve the reliability and accessibility of VLMs for agricultural diagnostics in developing countries. We present two key innovations:

\textbf{Prompt-Based Agricultural Expert Evaluation:} A scalable assessment protocol, used during the evaluation phase, that configures a language model to act as an expert plant pathologist. This enables automated, domain-aware scoring of generated captions without requiring constant access to human experts.

\textbf{Semantic Clustering for Self-Consistency:} A lightweight inference method that generates multiple candidate responses for an image and uses a cosine similarity-based consensus mechanism to select the most reliable and coherent diagnosis.

Our framework, applied to the PlantVillage dataset \cite{hughes2015open} using a fine-tuned 3B-parameter PaliGemma model, demonstrates improvements over standard decoding methods. When selecting the single most likely correct response, our semantic clustering method achieves a peak accuracy of 83.1\%, an improvement over the 77.5\% baseline accuracy of greedy decoding. Furthermore, our approach is more effective at surfacing a correct answer within a small set of candidates; the accuracy of finding a correct response within the top four semantic clusters reaches a peak of 94.0\%, outperforming the Greedy Top-4 selection method, which achieves 88.5\%.

\vspace{-2mm}
\section{Related Work}
\label{sec:related}

Agricultural image analysis has seen a progression from early methods using hand-crafted descriptors to deep CNNs, which achieved high accuracy on classification tasks. The PlantVillage dataset \cite{hughes2015open}, comprising over 54,000 images across 14 crop species and 39 disease categories, became a standard benchmark for these classification models \cite{mohanty2016using}. More recently, vision–language models (VLMs) have enabled the more ambitious goal of generating full-text captions that include symptom descriptions and treatment advice \cite{radford2021learning}. Domain-specific VLMs like WDLM for wheat rust \cite{zhang2024wheat} and AgriVLM for farm analytics \cite{yu2024agrivlm} have shown that curating image–text pairs boosts accuracy. However, these models typically rely on single-shot generation and may struggle to produce consistently reliable captions across the diverse range of crops and diseases found in PlantVillage. Our work addresses this by introducing a multi-response \emph{self-consistency} layer driven by a lightweight, 80MB embedding model, improving reliability without retraining the vision encoder or exceeding on-device resource budgets.

Reliability and efficiency remain open challenges, particularly for deployment in developing countries. Generic strategies that improve robustness in large language models—such as self-consistency voting \cite{wang2022self} and chain-of-thought reasoning \cite{wei2022chain}—are rarely ported to agricultural applications. While models like AgroGPT have explored synthetic data for instruction tuning, they still depend on a cloud-scale backbone unsuitable for edge deployment \cite{awais2025agrogpt}. Our framework adapts these ideas in a practical manner. We implement a self-consistency mechanism using a lightweight, pre-trained embedding model to perform cosine-consistency voting at inference time. This approach improves caption reliability without the high computational cost of repeated large-LM calls. For evaluation, we introduce a scalable protocol where an LLM prompted as a plant-pathology expert scores the final outputs.

\vspace{-2mm}
\section{Methodology}
\label{sec:method}

\subsection{Dataset Preparation and Caption Generation}
The foundation of our work is the publicly available PlantVillage dataset \cite{hughes2015open}, which provides a large collection of crop leaf images with corresponding disease labels. To train a vision-language model capable of generating detailed agricultural advice, we first converted the dataset's simple image-label pairs into rich, structured image-caption pairs.

We randomly selected 4,800 images from the PlantVillage dataset, spanning 14 crop types and 39 disease categories. The distribution reflects the diversity of agricultural challenges, with tomato comprising the largest subset (1,649 images) due to its eight distinct disease categories, followed by soybean (449 images), orange (459 images), grape (373 images), and corn (348 images). Crops with fewer disease categories have correspondingly smaller representation, such as raspberry (23 images) and strawberry (126 images), which have only healthy variants in our subset. The dataset encompasses a wide range of plant pathologies, from bacterial infections like tomato bacterial spot to fungal diseases such as grape black rot and corn common rust, providing comprehensive coverage for agricultural diagnostic training. For each image, we used its label (e.g., "Maize\_\_\_Cercospora\_leaf\_spot Gray\_leaf\_spot") to prompt a large language model (`gemini-2.5-pro`) \cite{gemini2025pushing}. The prompt instructed the model to act as an expert plant pathologist, analyze the visual information in the image, and generate a concise, five-part diagnostic report.\footnote{The full prompt text is available at \url{https://sigport.org/documents/datasetgenerationprompt}.} The required format specified a line each for: \textbf{(1)} disease name and infection level, \textbf{(2)} key visual symptoms, \textbf{(3)} a brief analysis, \textbf{(4)} specific fungicide recommendations, and \textbf{(5)} key prevention measures. This process resulted in a high-quality, synthetic dataset of 4,800 image-caption pairs, which served as the ground truth for fine-tuning and evaluation.

\begin{figure}[t]
    \centering
    \includegraphics[trim={0 0 0 0}, clip, width=0.33\columnwidth]{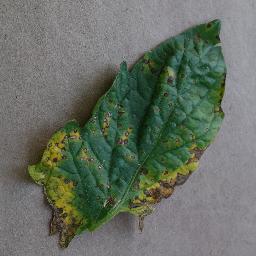}
    \caption{Image of a tomato leaf from PlantVillage dataset with its synthetic caption: "tomato early blight, infection level is moderate to severe; symptoms: dark brown circular spots, concentric rings, prominent yellowing, necrosis; analysis: established infection requires immediate fungicidal application; fungicides: chlorothalonil, mancozeb, azoxystrobin; prevention: crop rotation, resistant varieties, proper spacing, sanitation"}
    \label{fig:dataset_example}
\end{figure}

Figure~\ref{fig:dataset_example} shows an example of an image of a tomato leaf with early blight, paired with its detailed, synthetically generated caption.

\subsection{Vision-Language Model Fine-tuning}
We fine-tuned the 3B-parameter PaliGemma model \cite{beyer2024paligemma} using the JAX-based Big Vision framework. From our generated dataset, we used 4,000 image-caption pairs for training. To ensure computational efficiency, we employed a parameter-efficient fine-tuning (PEFT) strategy, exclusively updating the attention layers (`llm/layers/attn/`) of the language model component. The model was trained for 3 epochs using a batch size of 8 and a cosine decay learning rate schedule with a base learning rate of 0.03. The maximum sequence length was set to 128 tokens. This targeted fine-tuning adapted the model to the agricultural domain, enabling it to understand specific crop disease terminology and generate structured, multi-part captions from visual inputs.

\subsection{Multi-Response Generation for Evaluation}
The remaining 800 image-caption pairs from our dataset were reserved for evaluation. For each of these 800 images, we used the fine-tuned PaliGemma model to generate 21 distinct candidate outputs. One response was produced using greedy decoding, providing a baseline, while the other 20 were generated using temperature sampling ($t=1.0$) to create a diverse set of diagnostic possibilities. This collection of 16,800 generated responses (800 images x 21 responses) formed the basis for our self-consistency analysis and evaluation.

\begin{figure*}[t]
    \centering
    \includegraphics[trim={0 0 0 0}, clip, width=.9\textwidth]{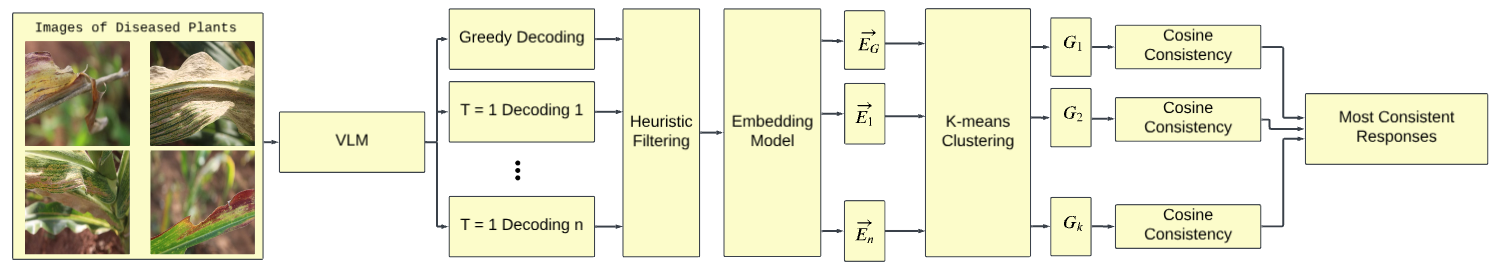}
    \caption{Farmer-captured crop images are analyzed by a fine-tuned vision-language model that produces candidate diagnoses; texts are encoded using a pre-trained embedding model, and cosine-consistency voting selects the most coherent diagnosis.}
    \label{fig:pipeline}
\end{figure*}

\subsection{Agricultural Image Processing Pipeline and Cosine-Consistency Voting}
Figure~\ref{fig:pipeline} illustrates our agricultural image processing framework. Our self-consistency mechanism leverages multiple diagnostic interpretations to identify the most reliable response through semantic consensus, following these steps:
\begin{enumerate}
    \item \textbf{Multi-Response Generation:} Generate 21 diverse candidate analyses using temperature sampling ($t=1.0$) and greedy decoding from the fine-tuned PaliGemma model.
\item \textbf{Heuristic Filtering:} Before embedding, we perform a lightweight filtering step to remove low-quality generations. This includes discarding responses containing error messages or inconsistent formatting (e.g., en/em dashes) and ensuring that the core diagnosis (the first word of the caption) matches the expected vegetable type. Specifically, the farmer visually confirms the crop type from the leaf image (e.g., identifying whether they are looking at a potato, tomato, or corn plant) and this user-provided crop identification is used to validate the generated responses. This manual check is a practical application of the \textbf{human-in-the-loop (HITL)} paradigm, where easily provided human context is used to guide and constrain the AI \cite{monarch2021human}. The need for human involvement to ensure reliability is critical in diagnostic systems, a principle that is also applied in other high-stakes domains like health informatics \cite{holzinger2016interactive}. In this step, the farmer provides a crucial data point for their specific context; by confirming the crop type through visual inspection of the leaf, they contribute to the quality and accuracy of the final output, a concept validated by data quality assessment frameworks in citizen science \cite{kosmala2016assessing}.
    \item \textbf{Domain-Aware Embedding:} Encode each of the filtered responses using the lightweight, pre-trained `all-MiniLM-L6-v2` sentence transformer model.
    \item \textbf{Consensus Calculation and Selection:} For each candidate response, we calculate its average cosine similarity to all other candidate responses in the set. The response with the highest average similarity score is selected as the final, most consistent output.
\end{enumerate}

This voting approach prioritizes responses that demonstrate agricultural consensus while filtering out outliers that may be linguistically coherent but agriculturally inaccurate.

\begin{table*}[t]
    \centering
    \caption{Accuracy comparison between the Greedy Top-K baseline and the Cosine-Consistency method. The 'Gens' column indicates the number of candidate responses available to the clustering algorithm. For the Greedy baseline, columns represent Top-K accuracy (K=1 to 4). For Cosine-Consistency, columns represent the accuracy of finding a correct response among the winners of K clusters.}
    \label{tab:results}
    \begin{tabular}{l|c|c|c|c|c|c|c|c}
        \toprule
        \multicolumn{1}{l|}{} & \multicolumn{4}{c|}{\textbf{Greedy}} & \multicolumn{4}{c}{\textbf{Cosine-Consistency}} \\
        \toprule
        \textbf{Gens} & \textbf{Top-1} & \textbf{Top-2} & \textbf{Top-3} & \textbf{Top-4} & \textbf{1-Cluster} & \textbf{2-Clusters} & \textbf{3-Clusters} & \textbf{4-Clusters} \\
        \toprule
        5  & 77.5\% & 83.4\% & 87.4\% & 88.5\% & 82.2\% & 88.3\% & 90.9\%  & 92.9\% \\
        10 & 77.5\% & 83.4\% & 87.4\% & 88.5\% & \textbf{83.1\%} & 88.3\% & 91.5\%  & 93.3\% \\
        15 & 77.5\% & 83.4\% & 87.4\% & 88.5\% & 81.9\% & 87.7\% & \textbf{93.1\%} & \textbf{94.0\%} \\
        20 & 77.5\% & 83.4\% & 87.4\% & 88.5\% & 83.0\% & \textbf{89.1\%} & 91.4\% & 93.8\% \\
        \bottomrule
        \end{tabular}
\end{table*}

\vspace{-2mm}
\section{Experimental Setup}
\label{sec:experiments}

\subsection{Dataset Partitioning}
From our synthetically generated dataset of 4,800 image-caption pairs, we partitioned the data for distinct purposes. A total of \textbf{4,000 samples} were used for fine-tuning the PaliGemma model. The remaining \textbf{800 samples} were reserved as a hold-out test set for evaluating the cosine-consistency algorithm. This test set was balanced across the 14 crop species and 39 disease categories to ensure a comprehensive assessment.

\subsection{Prompt-Based Scoring}
A central innovation for our evaluation is a prompt-based scoring framework that remedies the shortcomings of traditional text-overlap metrics.\footnote{The full prompt text for scoring is available at \url{https://sigport.org/documents/prompt-based-scoring}.} Through careful prompt engineering, we leverage the OpenAI o1-mini model \cite{o1-preview} and configure it as an expert plant pathologist for 14 crops and 39 diseases. The model is instructed to compare a generated description against a ground-truth original by analyzing five key sections (Diagnosis, Symptoms, Analysis, Fungicide, Prevention). It must determine if both descriptions refer to the same core disease and would lead to an equivalent management decision, while being flexible with synonymous terminology.

The similarity score follows a detailed rubric: \textbf{0.8--1.0} for the same core disease and treatment; \textbf{0.6--0.79} for the same disease with notable treatment differences; \textbf{0.4--0.59} for related conditions; and \textbf{0.0--0.39} for different diseases.

The model returns a Python dictionary containing the final score and a brief rationale. This framework is never invoked at inference time; it is used exclusively during our experiments to obtain reliable, domain-aware evaluation scores.

\subsection{Evaluation Methodology}
Our evaluation was conducted on the 800-sample test set. For each image, we generated 21 candidate responses from our fine-tuned PaliGemma model: one using greedy decoding and 20 using temperature sampling ($t=1.0$). A response is considered correct if its prompt-based score against the ground truth is $\geq 0.8$.

We compare two primary methods for selecting the best response from the candidates:
\begin{itemize}
    \item \textbf{Baseline (Top-K Selection):} This method assesses performance by selecting from the initial sequence of generated candidates without re-ranking. For a given $K$ (e.g., $K=4$), the first candidate is generated using greedy decoding, while the remaining $K-1$ candidates are generated via temperature sampling ($t=1.0$). The approach is considered successful if any of the first $K$ generated responses are correct. The Top-1 case for the baseline is the single response from greedy decoding.
    \item \textbf{Cosine-Consistency with Clustering:} This is our proposed self-consistency method. The generated responses are grouped using K-Means clustering (K=1 to 4). Within each cluster, a "winner" is selected based on which response has the highest average cosine similarity to all other members of that cluster. The approach is considered successful if any of the K cluster winners is correct.
\end{itemize}
\label{sec:dataset-partition}

\vspace{-2mm}
\section{Results and Analysis}
\label{sec:results}

Our experimental results, summarized in Table~\ref{tab:results}, demonstrate the significant benefits of using a cosine-consistency framework over a standard greedy baseline. The key findings are as follows:

\begin{itemize}
    \item \textbf{Superior Single-Choice Accuracy:} With 10 candidate generations, our 1-cluster cosine-consistency method achieves 83.1\% accuracy, outperforming the 77.5\% greedy baseline (Top-1) and demonstrating the value of semantic consensus.

    \item \textbf{Effectiveness of Clustering:} Accuracy increases with more clusters, rising from 81.9\% (1-cluster) to 94.0\% (4-clusters) with 15 generations. This shows clustering effectively surfaces a correct diagnosis within a small set of high-quality options.

    \item \textbf{Outperforming the Baseline:} At its peak (15 generations, 4 clusters), our method reaches 94.0\% accuracy, significantly outperforming the Greedy Top-4 baseline of 88.5\%. This confirms that semantic consensus is superior to simply selecting the first few generated outputs.
\end{itemize}

A key observation from Table~\ref{tab:results} is the trade-off between computational cost and accuracy. While increasing the number of candidate generations from 5 to 15 yields significant performance gains, the improvement diminishes when moving from 15 to 20 generations. This suggests that generating approximately 10-15 candidates offers a strong balance between diagnostic accuracy and the computational resources required, a critical consideration for deployment on resource-constrained mobile devices.

\vspace{-2mm}
\section{Discussion and Conclusions}
\label{sec:discussion}

This work introduces a practical framework for improving VLM-based agricultural diagnostics in developing countries. Our cosine-consistency algorithm achieves up to 94.0\% accuracy in identifying correct diagnoses within four candidate groups, providing users with a manageable set of high-quality options. The framework is computationally efficient, with strong performance using only 10-15 candidate generations, establishing a viable path toward production-ready diagnostic tools for agriculture.

\bibliographystyle{ieeetr}
\bibliography{main}

\end{document}